\documentclass[11pt,draftcls,onecolumn]{IEEEtran}

\usepackage{epsfig}
\usepackage{graphicx}
\usepackage{amsmath}
\usepackage{amssymb}
\usepackage{amsfonts}
\usepackage{subfigure}
\usepackage{txfonts}
\usepackage{comment}
\usepackage{color}
\usepackage[ruled,linesnumbered]{algorithm2e}
\usepackage[nospace,noadjust]{cite}

\usepackage[pagebackref=true,
breaklinks=true,
letterpaper=true,
dvipdfm,
colorlinks=true,
linkcolor=red,
hyperindex,
citecolor=green,
pdfstartview=FitH,
plainpages=false,
bookmarks=true
]{hyperref}

\ifCLASSINFOpdf
\else
\fi

\begin{document}
%
\title{Visual Vocabulary Learning and Its Application to 3D and Mobile Visual Search}

\author{
Liujuan Cao,~\IEEEmembership{Student Member,~IEEE}
\thanks{Liujuan Cao is with the School of Computer Science, Harbin Engineering University, 150001, P. R. China,
e-mail: caoliujuan@hrbeu.edu.cn.}
}

\markboth{Research Report}%
{Shell \MakeLowercase{\textit{et al.}}: Bare Demo of IEEEtran.cls for Journals}

\maketitle

\begin{abstract}
In this technical report, we review related works and recent trends in visual vocabulary based web image search, object recognition, mobile visual search, and 3D object retrieval. Especial focuses would be also given for the recent trends in supervised/unsupervised vocabulary optimization, compact descriptor for visual search, as well as in multi-view based 3D object representation.
\end{abstract}


\IEEEpeerreviewmaketitle

\section{Introduction}
\label{sec:introduction}
\IEEEPARstart{R}{ecently}, local feature representations \cite{Lowe04}\cite{Mikolajczyk05}\cite{Mikolajczyk06} are very popular in computer vision research, with extensive applications in near duplicate visual search, mobile visual search, video copy detection, image annotation, and web-scale image retrieval.
Generally speaking, the state-of-the-art visual search systems are built on the so-called visual vocabulary model, that is, (hierarchical) quantization of local feature spaces with inverted indexing speed up
\cite{Sivic03}\cite{Nister06}\cite{Philbin07}\cite{Schindler07}\cite{JiCVPR09}\cite{JiCVPR10}\cite{JiIEEEMM11}.
In this scenario, the local features \cite{Lowe04}\cite{Mikolajczyk05} extracted from reference images are quantized into a set of visual words, each with an indexing file.
Each reference image is then represented as a Bag-of-Words histogram and is inversely indexed into words that contain local features extracted from this image.
This Bag-of-Words representation offers sufficient robustness against photographing variances in occlusions, viewpoints, illuminations, scales and backgrounds.
Subsequently, the image search problem is reformulated from a document retrieval perspective, from which perspective many successful techniques such as TF-IDF \cite{Salton88}, pLSA \cite{Hofmann01} and LDA \cite{Blei03} can be directly used.

In general, both visual vocabulary and hashing based search techniques can be categorized into the approximate visual search techniques, which has been well exploited in the recent literature to handle the search deficiency in large image collections, \emph{e.g.} Vocabulary Tree \cite{Nister06}, Approximate K-means \cite{Philbin07}, Hamming Embedding \cite{Jegou08}, Locality Sensitive Hashing \cite{Grauman09} and their variances \cite{Jurie05}\cite{Yang07}\cite{Philbin07}\cite{Jegou10}\cite{JiCVPR09}\cite{JiTOMCCAP12b}.
In a typical setting, the visual vocabulary based search system works in a client-server paradigm:
The client end (\emph{e.g.} a mobile device or a web interface) sends a query image to the server. Or alternatively, in some recent mobile visual search systems \cite{Chen09}\cite{Chandrasekhar09}\cite{JiIJCV12}\cite{JiIJCAI11}\cite{JiMM11}\cite{JiICASSP11}, sending compact visual descriptors extracted from this query image can further reduce the wireless communication latency.

The server end searches similar reference images in its reference data set following three consecutive phases as:

\begin{itemize}
\item Extracting local features from the query image (if the server delivers compact visual descriptors instead of the query image, this step is skipped);
\item Quantizing these local features into a Bag-of-Words histogram using the vocabulary;
\item Ranking similar images in the inverted indexing files of all non-empty words, so as to avoid the linear scanning of all reference images in the similarity ranking.
\end{itemize}

In this technical report, we review related works and recent trends in visual vocabulary based web image search, object recognition, mobile visual search, and 3D object retrieval. Especial focuses would be also given for the recent trends in supervised/unsupervised vocabulary optimization, compact descriptor for visual search, as well as in multi-view based 3D object representation.

\section{Visual Vocabulary Construction}
In this section, we first review the recent advances in visual vocabulary construction.
Typically speaking, building visual vocabulary usually resorts to unsupervised vector quantization, which subdivides the local feature space into discrete regions each corresponds to a visual word.
An image is represented as a Bag-of-Words (BoW) histogram, where each word bin counts how many local features of this image fall in the corresponding feature space partition of this word.
To this end, many vector quantization schemes are proposed to build visual vocabulary,
such as K-means \cite{Sivic03}, Hierarchical K-means (Vocabulary Tree) \cite{Nister06}, Approximate K-means \cite{Philbin07},
and their variances \cite{Jurie05}\cite{Yang07}\cite{Philbin07}\cite{Jegou10}\cite{JiICME08b}\cite{JiCVPR09}\cite{JiCVPR12b}\cite{JiCVPR12c}.
Meanwhile, hashing local features into a discrete set of bins and indexed subsequently is an alternative choice, for which methods like Locality Sensitive Hashing (LSH) \cite{Gionis99}, Kernalized LSH \cite{Grauman09}, Spectral Hashing \cite{Weiss08} and its variances \cite{JiCVPR12a}\cite{JiTOMCCAP12b} are also exploited in the literature.
The visual word uncertainty and ambiguity are also investigated in \cite{Jegou08}\cite{Philbin07}\cite{Jiang07}\cite{Gemert09}, using methods such as
Hamming Embedding \cite{Jegou08}, Soft Assignments \cite{Philbin07} and Kernelized Codebook \cite{Gemert09}.
Some other related directions include optimizing the initial inputs of visual vocabulary construction, such as learning a better local descriptor detector as in \cite{JiTIST12}, coming up with a better similarity metric, such as learning an optimal hashing based distance matching as in \cite{JiPR11} for human action search  \cite{JiMM08}, incorporating Bayesian reasoning into the similarity calculation \cite{JiIJICIC08}, as well as distribute the visual vocabulary model and its inverted indexing structure into multiple machines \cite{JiTMM12}.

Stepping forward from unsupervised vector quantization, semantic or category labels are also exploited \cite{Moosmann06}\cite{Mairal08}\cite{Lazebnik09}\cite{JiCVPR10}\cite{JiTIP12a} to supervise the vocabulary construction, which learns the vocabulary to be more suitable for the subsequent classifier training, \emph{e.g.}, the images in the same category are more likely to produce similar BoW histograms and vice versa.
In terms of functionality, works in \cite{Mairal08}\cite{Lazebnik09}\cite{Moosmann06} exploits the learning-based codebook constructions.
For instance, Mairal et al. \cite{Mairal08} built a supervised vocabulary by learning discriminative and sparse coding models for object categorization.
Lazebnik et al. \cite{Lazebnik09} proposed to construct supervised codebooks by minimizing mutual information lost to index fully labeled data.
Moosmann et al. \cite{Moosmann06} proposed an ERC-Forest to consider semantic labels as stopping tests in
building supervised indexing trees.
Another group of related works~\cite{Perronnin06}\cite{Zhang07}\cite{Liu09Learning} refines (merges
or splits) the initial codewords to build class(image)-specific vocabularies for categorization. Although working well for limited-number categories, these approaches cannot be scaled up to generalized
scenarios with numerous and semantically correlative categories.
Similar works can also be referred to the Learning Vector Quantization \cite{Kohonen86}\cite{Kohonen00}\cite{Rao96} in data
compression, which adopted self-organizing maps \cite{Kohonen00} or regression lost minimization \cite{Rao96} to build codebook that minimizes training data distortions after compression.
From the supervised learning point of view, works in topic decompositions (pLSA \cite{Li07} or LDA \cite{Bosch08}\cite{JiCVPR12b}) can be also treated as supervised codebook refinement, which typically resorts to learning a topical-level representation instead of the word-level representation.
It is worth to note that,  by exploiting semantics learning in visual representation stage, this kind of works differs from works that adopt semantic learning to refine the subsequent recognition stages \cite{Duygulu02}\cite{Marszalek07}\cite{Liu09}, e.g. classifiers based on machine translation \cite{Duygulu02} or semantic hierarchy \cite{Marszalek07}.
On the other hand, optimizing the visual vocabulary can be also benefit the related tasks such as image annotation, relevance feedback learning, video location search and text detection, as shown in \cite{JiMM09c}\cite{JiICASSP09}\cite{JiVCIP08}\cite{JiICME08a}\cite{JiICME08c}\cite{JiMIR08a}\cite{JiMIR08b}.

\section{Landmark Search and Recognition}

A widely exploit scenario in visual vocabulary model comes from landmark search and recognition, where the near-duplicate changes in terms of different viewing angles, occlusions, appearance changes, as well as perspective changes fits the merits of visual  vocabulary model well, with lots of applications such as mobile localization and advertisement recommendation \cite{JiMM09c}.
Recently, scalable near-duplicate image retrieval \cite{Nister06,Philbin07,JiICME08b,JiCVPR09,JiICASSP11,JiIJCV12} has been largely addressed by promising visual vocabulary models as well as inverted indexing techniques.
The representative approaches for vocabulary construction include K Means clustering \cite{Sivic03}, Vocabulary Tree \cite{Nister06}, and Approximate K-Means \cite{Philbin08} et al.
We organize this topics from the following two perspectives based upon the problem scale, named city-scale landmark search and worldwide landmark search:

\subsection{City-Scale Landmark Search}
Towards city-scale landmark search and recognition,
Schindler et al. \cite{Schindler07} presented a location recognition system through
geo-tagged video streams with multiple path search in the vocabulary tree.
Eade et al. \cite{Eade08} also adopted a vocabulary tree for real-time loop closing based on SIFT-like descriptors.
Our previous works in \cite{JiCVPR09} proposed a density-based metric learning to optimize the hierarchical structure of vocabulary tree \cite{Nister06} for street view location recognition.
%
%
Yeh et al. \cite{Yeh07} further adopted a hybrid color histogram to compensate the feature based ranking in mobile based location recognition applications.
Cristani et al. \cite{Cristani08} learnt a global-to-local image matching for location recognition.
And their consecutive work in \cite{Crandall09} identified landmark buildings based on image data,
metadata, and other photos taken within a consecutive 15-minute window.
In addition, Irschara et al. \cite{Irschara09} further leverage structure-from-motion (SFM) to build 3D scene models for street views,
combined with vocabulary tree for simultaneously scene modeling and location recognition.
Xiao et al. \cite{Xiao08} proposed to combine bag-of-features with simultaneous localization and mapping (SLAM) to further improve the recognition precision.
The quantization issues in visual vocabulary are recently also well addressed to fit the city-scale landmark search scenario, such as the works in \cite{JiCVPR09}\cite{JiIEEEMM11}.
Incrementally vocabulary indexing is also explored in \cite{JiICME08b} to maintain a landmark search system in a time varying database.

\subsection{Worldwide Landmark Search}
Towards worldwide landmark search and recognition, the IM2GPS system \cite{Hays08}
inferred possible location distributions of a given query by visual matching in a worldwide, geo-tagged landmark dataset.
As a consecutive work, Kalogerakis et al. further \cite{Kalogerakis09} demonstrated how to combine single image matching with sequential data to improve matching accuracy.
Zheng et al. \cite{Zheng09} developed a worldwide landmark recognition system,
which used a predefined landmark list to query online image search engines to selected candidate images, followed by re-clustering and pruning to
locate the final landmark location.
Recent works also proposed to mine representative landmarks at a worldwide scale, such as the sparse representation based landmark mining approach in \cite{JiTOMCCAP11}\cite{JiMM09a}.

\section{3D Object Retrieval and Recognition}
Recently, extensive research efforts \cite{Paquet2000,Gao2010MMBNT, Vranic2003,Gao2010MM1,Xiao2011a,Leibe2003,Xiao2011b} have been dedicated to 3D object retrieval and recognition, and how to effectively and efficiently search 3D objects is an important topic in multimedia research. Early methods \cite{Ip2002,Leng2008,Funkhouser2003,Tangelder2003,Gao2010PR} mainly focus on model-based method, while the view-based methods \cite{Thomas2006,Gao2010NC,Savarese2007,Gao2011SPIC,Su2009,Gao2012TIP1,Ohkita2012,Gao2012TIP3,Sun2009,Gao2012INS,Gao2011TMM} have attracted much attention nowadays. This is due to the fact that view-based methods \cite{Gao2012TIP2,Ansary2007TMM,Gao2011MM} are with the highly discriminative property for object representation and visual analysis \cite{Gao2009MTA,Gao2009MM,Gao2009CIVR,Gao2008TCE,Gao2008ICIP,Wang2007MM,Wang2007MMW}  also plays an important role in multimedia applications. For view-based 3D object retrieval, the visual words have been investigated for 3D object representation.

In these view-based 3D object retrieval methods, generally the SIFT features are extracted from all selected views, and a visual word dictionary is learnt. Then a bag-of-visual-words description is generated for 3D object representation. The matching between two 3D objects is conducted based on this representation.

The major advantages by using visual words description in 3D object retrieval and recognition are two-fold.
\begin{enumerate}
  \item The visual words description is effective on image representation, which can be disciminative for the description of different classes of objects.
  \item The visual words can be extracted easily, and it is robust to object scaling and rotating.
\end{enumerate}

Furuya and Ohbuchi \cite{Furuya2008a} first proposed to extract SIFT features for views of 3D objects. In this method, each 3D object is rendered into a group of depth images, and the SIFT features are extracted from these images. This method uses the bag-of-features approach to integrate the local features into a feature vector for each model. Then the matching of these two feature vectors determines the distance between the two 3D objects. Ohbuchi et al. \cite{Ohbuchi2008c} further proposed to employ Kullback-Leibler divergence (KLD) to calculate the distance between two bag-of-visual-feature based 3D objects. Osada et al. \cite{Osada2008a} employed the bag-of-visual-feature method to SHREC'08 CAD model track task.  Ohbuchi and Shimizu \cite{Ohbuchi2008b} employed the semi-supervised manifold learning method for object class recognition.  The proposed method projects the original feature space onto a lower dimensional manifold. Then the relevance feedback information is employed to capture the semantic class information by using the manifold ranking algorithm. Though the bag-of-visual-feature description is effective on 3D object retrieval, the computational cost is high. Ohbuchi and Furuya \cite{Ohbuchi2008a} further accelerated the method by using a Graphics Processing Unit. A bag-of-region-words method \cite{Gao2010MM2} is introduced to extract visual features in the region level. This method first gridly selects points in each image and the local SIFT features are extracted for these points. Then each feature is encoded into a visual word with a pre-trained visual vocabulary. In this step, each view is split into a set of regions, and each region is represented by a bag-of-visual-words feature vector. All the obtained regions are further grouped into clusters based on the bag-of-visual-words feature, and one feature is chosen as the representative one from each cluster with corresponding weight.  The Earth Movers Distance is used to measure the distance between two 3D objects.

Furuya and Ohbuchi \cite{Ohbuchi2009a} proposed to employ dense sampling to extract feature points in the views of 3D objects,  and then the SIFT feature is extracted from each point. These visual features are further clustered into groups to generate the visual vocabulary. Then the feature histogram is generated to calculate the 3D object distance. This method has been further extended \cite{Ohbuchi2009a,Ohbuchi2010b} to deal with large scale data. A distance metric learning method \cite{Ohbuchi2010a} is proposed to learn the distance metric for matching of 3D models. Endoh et al. \cite{Endoh2012} introduced  to conduct learning on the manifold structure of 3D models by using clustering-based training sample reduction. Kawamura et al. \cite{Kawamura2012} further employed the geometrical feature to improve the feature-based method. 

\section{Mobile Visual Search}

\subsection{The Need of Compact Descriptor}
There are many evidences support the usage of compact descriptors for mobile visual search and augmented reality applications:

\begin{itemize}
\item Firstly, it remains a long way to provide a stable and high-speed (3G) wireless coverage everywhere, especially for those touristic landmarks that are often far away from urban areas or for developing countries, e.g., Lhasa, Tibet in our experiments. So it is unrealistic to guarantee the bandwidth is good enough to reliably and fast send a query photo. In particular, the recently established MPEG Ad Hoc Group \textbf{CDVS} is bring together the academia and industry practitioners to explore the next MPEG standard of \textbf{C}ompact \textbf{D}escriptor for \textbf{V}isual \textbf{S}earch.
\item Secondly, from the server perspective, the network capability of receiving a batch of entire photo queries is by no doubt limited for a more powerful cloud platform that may handle intensive search at the server end. From the industry practice, a clear fact is that receiving multiple query photos is much more challenging than receiving texts in the state-of-the-art search engines. More importantly, with compact upstream queries, more bandwidth can be saved up to downstream return the actually valuable searched information (in rich forms of text, images and video). That is one of the reasons why many internet service providers often set a smaller uplink bandwidth to save bandwidth for fast browsing.
\item Finally, sending large amount of data via 3G wireless definitely causes serious battery energy consumption. Empirical evidence shows that compressing the query photo into a compact signature and sending the signature through the mobile is much more power saving.
\end{itemize}

In summary, the promising research efforts in compact visual descriptors are bringing great benefits in lightening the battery consumption, the cost of bandwidth and memory , which undoubtedly contribute to efficient and effective visual query delivery in mobile visual search, especially in the scenarios of video rate reality augmentation.

\subsection{The State of The Arts}

The ever growing computational power motivates the research efforts to extract visual descriptors directly  on a mobile device \cite{Chen09,Chen09Another,Chandrasekhar09,Chandrasekhar09VCIP,Makar09,JiMMS09,JiIJCAI11,JiICIP11a,JiICIP11b}.
Instead of sending an entire photo, sending such descriptors are compact enough to enable the low bit rate search.
Comparing with the previous works in low dimensional local descriptors
such as PCA-SIFT \cite{Ke04}, GLOH \cite{Mikolajczyk05}, SURF \cite{Bay06}, and MSR descriptors \cite{Hua07},
works  in \cite{Chen09,Chen09Another,Chandrasekhar09,Chandrasekhar09VCIP,Makar09} target at intensive compactness as well as efficient extraction in a standard mobile end.
They are expected to work well in mobile visual search scenarios.

Coming with the ever growing computational power in the mobile devices,
recent works have proposed to directly extract compact visual descriptors on the mobile devices \cite{Chen09,Chen09Another,Chandrasekhar09,Chandrasekhar09VCIP,Makar09}.
Instead of sending the entire query, such descriptors are transmitted to enable a low bit rate search.
Comparing with the previous works in low dimensional local descriptors
such as PCA-SIFT \cite{Ke04}, GLOH \cite{Mikolajczyk05}, SURF \cite{Bay06}, and MSR descriptors \cite{Hua07},
works in \cite{Chen09,Chen09Another,Chandrasekhar09,Chandrasekhar09VCIP,Makar09} target at very extreme compression rates
as well as efficient online extraction in the mobile end.
Consequently, recent works in \cite{Chen09,Chen09Another,Chandrasekhar09,Chandrasekhar09VCIP,Makar09}
have focused on more compact descriptors specialized for the mobile visual search:

Towards compact local visual descriptors, Chandrasekhar et al. proposed a Compressed Histogram of Gradient (CHoG) \cite{Chandrasekhar09},
which are further compressed by both Huffman Tree and Gagie Tree to reduce the size of each descriptor to approximate 50 bits.
Works in \cite{Chandrasekhar09VCIP} employ Karhunen-Loeve transform to compress the SIFT descriptor,
producing approximate 2 bits per SIFT dimension (128 dimensions in total).
Tsai et al. \cite{Tsai10} proposed to transmit the spatial layouts of interest points to improve the precision of feature matching.
Comparing with sending an entire query photo, sending above compact descriptors are much more efficient \cite{Chandrasekhar10Review}.
For instance, CHoG typically outputs only 50 bits per local feature.
When 1,000 interest points are extracted per query (following the popular detector setting \cite{Mikolajczyk06}),
the data amount to transmit is only 8KB, much less than the entire query photo (typically over 20KB with JPEG compression).

Chen et al. \cite{Chen09} stepped forward to send the bag-of-features histogram \cite{Chen09,Chen09Another} instead,
which encodes the position difference of non-zero bins to yield approximate 2KB per query photo using a one million vocabulary.
It largely outperforms directly sending the compact local descriptors (more than 5KB in reported works).
Their successive work in \cite{Chen09Another} further compressed the inverted indexing structure of visual vocabulary \cite{Nister06}
with arithmetic coding to reduce the memory and storage cost to maintain the visual search system in server(s).
A recent group of representative works come from the endeavors of Ji et al. in compression the visual vocabulary based descriptor representation directly on the mobile end \cite{JiIJCV12}\cite{JiCVPR12b}\cite{JiIJCAI11}\cite{JiMM11}\cite{JiICASSP11}\cite{JiICIP11b}.

Beyond the context of mobile visual search, compact image signatures are recently investigated in
\cite{Weiss09}\cite{Jegou10}\cite{Jegou10PAMI}\cite{JiTIP12a}.
For instance, Jegou et al. proposed a product quantization scheme \cite{Jegou10PAMI} to learn a compact image descriptor that approximates
the square distance of original Bag-of-Words histograms.
The same authors also proposed a miniBOF feature \cite{Jegou09} by packing the bag-of-features.
Their recent work in \cite{Jegou10} further aggregated local descriptors with PCA and locality sensitive hashing to produce a compact descriptor of approximate 32 bits in length..
Weiss et al. \cite{Weiss09} used spectral hashing to compress GIST descriptor \cite{Torralba08} into tens of bits.
Wang et al. \cite{Wang2010_new} proposed a locality-constrained linear coding (LLC) scheme over the Bag-of-Words histogram to improve the spatial pyramid matching.
In multi-view coding, Yeo et al. \cite{Yeo08} proposed a rate-efficient correspondence learning scheme to randomly project descriptors to build a minHashing code.

\ifCLASSOPTIONcaptionsoff
  \newpage
\fi

\bibliographystyle{ieee}

%

\end{document}